\documentclass[conference,10pt]{IEEEtran}
\IEEEoverridecommandlockouts
\usepackage{cite}
\usepackage{amsmath,amssymb,amsfonts}
\usepackage{graphicx}
\usepackage{textcomp}
\usepackage{xcolor}
\usepackage{booktabs} 
\usepackage{float}
\usepackage{multirow}
\usepackage{enumitem}
\usepackage{mathtools}
\usepackage[letterpaper]{geometry}
\usepackage[linesnumbered,ruled]{algorithm2e}
\usepackage{amsmath}
\usepackage{tikz}
\newgeometry{left=1.5cm,right=1.5cm,bottom=1.7cm,top=1.7cm}
\renewcommand{\baselinestretch}{0.985}
\usepackage{array}

\newcommand*\circled[1]{\tikz[baseline=(char.base)]{
            \node[shape=circle,fill,inner sep=1pt] (char) {\textcolor{white}{#1}};}}

\def\BibTeX{{\rm B\kern-.05em{\sc i\kern-.025em b}\kern-.08em
    T\kern-.1667em\lower.7ex\hbox{E}\kern-.125emX}}
\begin{document}

\title{
\vspace{-25pt}
\small © 2025 IEEE. This is the author’s version of the work. The definitive Version of Record was Published in the\\
Design Automation Conference (DAC), San Francisco, United States, June 2025.\\
\rule{\linewidth}{0.4pt} \\[5pt]
\huge ESM: A Framework for Building Effective Surrogate Models for Hardware-Aware Neural Architecture Search \vspace{-10pt}
}


\author{\IEEEauthorblockN{Azaz-Ur-Rehman Nasir$^*$, Samroz Ahmad Shoaib$^*$, Muhammad Abdullah Hanif$^*$, Muhammad Shafique}
\IEEEauthorblockA{\textit{eBRAIN Lab, Division of Engineering, New York University Abu Dhabi, UAE} \\
\{an3525, ss14256, mh6117, muhammad.shafique\}@nyu.edu}\vspace{-25pt}}

\maketitle
\def\thefootnote{*}\footnotetext{These authors contributed equally to this work}\def\thefootnote{\arabic{footnote}}

\begin{abstract}
Hardware-aware Neural Architecture Search (NAS) is one of the most promising techniques for designing efficient Deep Neural Networks (DNNs) for resource-constrained devices. 
Surrogate models play a crucial role in hardware-aware NAS as they enable efficient prediction of performance characteristics (e.g., inference latency and energy consumption) of different candidate models on the target hardware device. 
In this paper, we focus on building hardware-aware latency prediction models. 
We study different types of surrogate models and highlight their strengths and weaknesses. 
We perform a systematic analysis to understand the impact of different factors that can influence the prediction accuracy of these models, aiming to assess the importance of each stage involved in the model designing process and identify methods and policies necessary for designing/training an effective estimation model, specifically for GPU-powered devices. 
Based on the insights gained from the analysis, we present a holistic framework that enables reliable dataset generation and efficient model generation, considering the overall costs of different stages of the model generation pipeline.

\end{abstract}


\section{Introduction and Motivation}
\label{Sec1:Introduction}

\begin{figure}[h]
    \centering
    \includegraphics[width=1\linewidth]{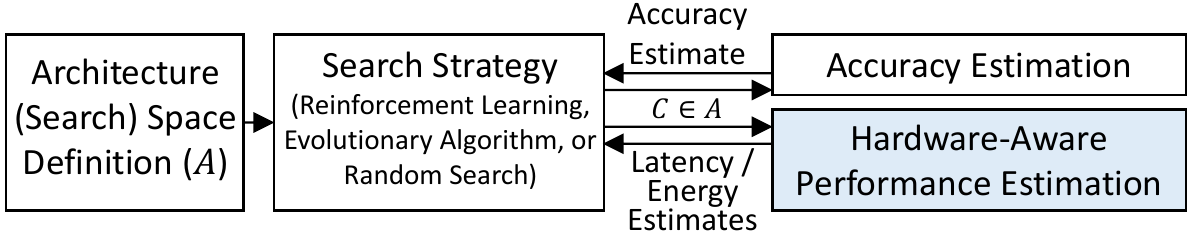}
    \vspace{-20pt}
    \caption{A general overview of hardware-aware NAS which highlights the importance of hardware-aware performance (latency or energy) estimation for guiding the architecture search strategy.\vspace{-10pt}}
    \label{fig:NAS_Overview_1}
\end{figure}

Deep Neural Networks (DNNs) are nowadays being used to serve a wide variety of tasks in different domains such as computer vision~\cite{voulodimos2018deep} and Natural Language Processing (NLP)~\cite{otter2020survey}\cite{zhao2023survey}. 
Moreover, the diverse set of devices used for DNN deployment in real-world application (e.g., embedded CPU and GPU devices, as well as specialized hardware accelerators~\cite{chen2020survey}) brings a unique challenge, as DNNs result in different latency and energy numbers on different devices because of the differences in their hardware architecture and capabilities~\cite{cai2019once}. 
Neural Architecture Search (NAS), specifically, hardware-aware NAS (HW-NAS), techniques have been proposed to automatically search for DNNs based on the user-defined performance and hardware constraints~\cite{benmeziane2021hardware}. A generic overview of hardware-aware NAS techniques is presented in Fig.~\ref{fig:NAS_Overview_1}. 

Various HW-NAS techniques have been proposed in the literature. One prominent set of techniques focuses on simultaneous exploration and training of candidate models to converge to a suitable solution. 
In contrast, methods like Once-for-All (OFA)~\cite{cai2019once} adopt a decoupled strategy, where a supernet is first trained, followed by exploration of DNN configurations based on user-defined constraints, without involving additional training. 
The core benefit of the latter category lies in fast exploration, as it eliminates model training iterations. 
Building on this advantage, our work focuses on improving the exploration phase of OFA-based techniques to efficiently and accurately identify suitable models for deployment on target hardware, while satisfying all user-defined performance constraints.

\begin{figure}[h]
    \centering
    \includegraphics[width=0.98\linewidth]{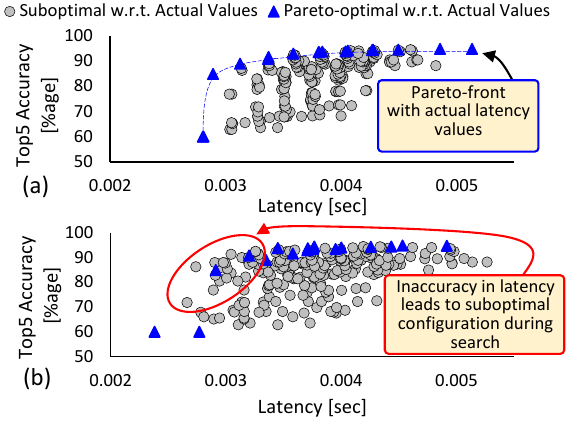}
    \vspace{-15pt}
    \caption{Impact of inaccuracy in latency prediction on neural architecture search results. (a) illustrates the top-5 accuracy vs. latency plot of 243 different ResNet architectures generated by varying the depth values of the pre-trained ResNet50 supernet made available by Cai et al.~\cite{cai2019once}. The accuracy numbers are generated by evaluating each candidate model on the ImageNet validation set, and the latency numbers are generated by running the model on an RTX4090 GPU. (b) illustrates how the actual Pareto-optimal points can potentially move in the architecture space when the latency prediction is slightly inaccurate.\vspace{-20pt}}
    \label{fig:motivation1}
\end{figure}

Since deploying each candidate model on hardware to evaluate its performance is time-consuming, HW-NAS techniques utilize surrogate models to efficiently estimate the performance or latency of DNNs on the target hardware~\cite{yang2018netadapt}\cite{cai2018proxylessnas}\cite{wu2019fbnet}\cite{wang2020apq}\cite{wang2020hat}. Fig.~\ref{fig:motivation1} highlights the importance of using a precise latency prediction model to converge to models that lie on or near the actual Pareto front. 
Various techniques have been proposed for building surrogate models for HW-NAS. 
Earlier methods relied on proxy metrics (e.g., FLOPs) to estimate DNN performance~\cite{tan2019efficientnet}\cite{zhang2018shufflenet}. However, such metrics are hardware-agnostic and fail to accurately reflect a model's true performance or latency on the target hardware. 
To address this limitation, lookup table and ML-based techniques have been proposed for latency prediction. 
Lookup table-based techniques use an additive model where latency of each individual layer is taken from the lookup table (defined through profiling) and then the latencies of all the layers are accumulated to produce the latency of the complete model. 
These techniques result in suboptimal results as they lack in capturing the complex interactions between layers. 
The ML-based techniques encode the model architecture into an intermediate representation, which serves as input for performance predictors. 
The most common architecture encoding types include one-hot~\cite{mauch2020efficient}\cite{dai2021fbnetv3}, and feature representation encodings~\cite{hassantabar2022curious}\cite{li2022inference}\cite{mendoza2020predicting}. 
These are used with models such as multilayer perceptron (MLP), linear regression, decision tree regression, and boosted decision tree regression~\cite{dai2021fbnetv3}\cite{mendoza2020predicting}\cite{hassantabar2022curious}\cite{li2022inference}. 
Other techniques leverage the graphical structure of DNNs to utilize Graph Convolutional Networks (GCNs) for building performance predictors~\cite{mauch2020efficient}\cite{ning2020generic}. 
However, these ML models lack in providing accurate latency estimates as will be demonstrated in the following motivational case study.

\subsection{Motivational Case Study}

\begin{figure}
    \centering
    \includegraphics[width=1\linewidth]{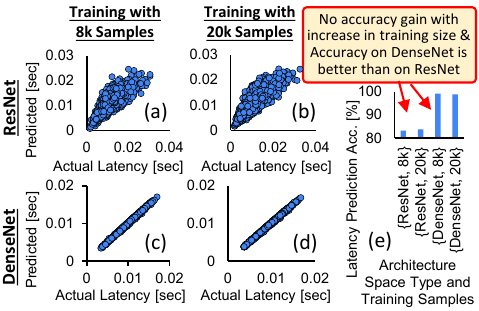}
    \vspace{-20pt}
    \caption{Actual vs. predicted latency for 4k ResNet model configurations using (a) a predictor trained with 8,000 samples and (b) a predictor trained with 20k samples. 
    (c) and (d) present the results for DenseNet configurations. (e) Comparison of the average accuracy of predictors trained with 8,000 samples vs. those trained with 20,000 samples for the two model types.\vspace{-10pt}}
    \label{fig:motivation2}
\end{figure}

In this section, we highlight the impact of increasing the training set size on the accuracy of latency prediction models. 
For this analysis, we considered two different supernet architecture types: ResNet and DenseNet architectures. 
The configurations for both are presented in the experimental setup section \textcolor{black}{(III.A)}. Note that the architecture space considered for ResNet supernet is much larger than that considered for DenseNet. 
For prediction model, we used MLP together with summary feature representation encoding from~\cite{wang2020hat}. 
For each architecture type, we randomly sampled 24000 samples from the architecture space and executed them on an RTX4090 machine to gather architecture-latency pairs for predictor training and evaluation. 
We used 4000 samples for testing and 8000 or 20000 for training. 
Fig.~\ref{fig:motivation2}(e) shows that, for both the architecture types, increasing the training set size from 8000 architecture samples to 20000 samples does not result in any significant improvement in the prediction accuracy. 
Moreover, as the architecture space considered for DenseNet supernet is much smaller than that of ResNet, the predictor results in significantly higher accuracy for DenseNet. 
We argue that the lower accuracy in the case of ResNet model is due to the overlapping representation of configurations in the representation space, and thus, \textit{there is a need to explore alternative encoding schemes that can result in better accuracy for larger OFA architecture spaces.} 

\begin{figure}
    \centering
    \includegraphics[width=1\linewidth]{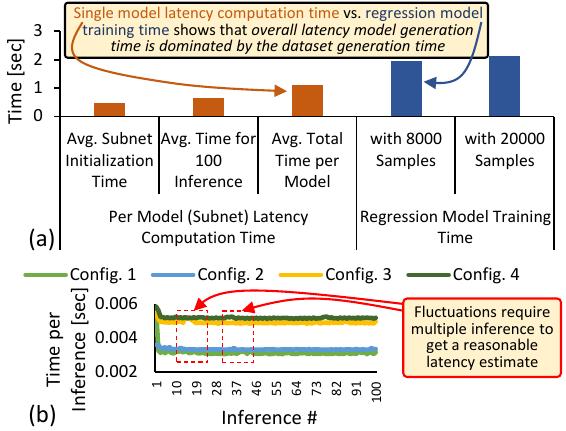}
    \vspace{-20pt}
    \caption{(a) Comparison between the time required to accurately compute the latency of a model derived from the pre-trained ResNet50 supernet by Cai et al.~\cite{cai2019once} on an RTX4090 GPU and the training time for a latency predictor. On average, training a regression model of reasonable size on an RTX4090 GPU takes approximately the same amount of time as computing the latency of a model on the target hardware. (b) Latency values across inferences for different architectural configurations, illustrating the need for multiple iterations to obtain a reliable value.\vspace{-10pt}}
    \label{fig:motivation3}
\end{figure}

In addition to the above, we compare the time required to accurately compute the latency of a sample architecture on an RTX4090 GPU with the time needed to train a latency predictor. 
Fig.\ref{fig:motivation3}(a) shows that the average latency measurement time per model is comparable to the time required to train a latency predictor with over 8,000 samples. 
Note that each latency measurement involves multiple inferences to account for fluctuations, as illustrated in Fig.\ref{fig:motivation3}(b). 
\textit{On one hand, this emphasizes the high cost of data acquisition, while on the other hand, it highlights the potential for an iterative methodology—comprising training, testing, and dataset extension if necessary—to ensure early exit and minimal overall model generation cost.}







\subsection{Our Novel Contributions}
To address the above challenges, in this paper, we make the following key contributions. 
\begin{enumerate}
    \item We propose ESM, a framework for building effective latency prediction models for HW-NAS algorithms. 
    ESM targets layer/block-wise architecture spaces built on a fixed macro-architecture. The framework adopts a \textit{train-evaluate-extend} approach, allowing it to efficiently converge on an accurate model that meets user-defined accuracy constraints while minimizing dataset collection costs.    
    \item We propose multiple different architecture encoding schemes specialized for building accurate latency prediction models for layer/block-wise operation space built on a fixed macro-architecture. Our encoding schemes provide a balance between accuracy and surrogate model complexity, without significantly increasing the need for larger training datasets to achieve high accuracy. This is achieved by enhancing the representation capability of the encoding while minimizing the risk of increased sparsity, which could otherwise require larger datasets. 
    \item Our results on a variety of models and devices demonstrate the effectiveness of our proposed encoding schemes over state-of-the-art encoding schemes, specifically for highly diverse architecture search spaces based on fixed macro-architectures. 
\end{enumerate}

\section{Methodology}
\begin{figure*}
    \centering
    \includegraphics[width=1\linewidth]{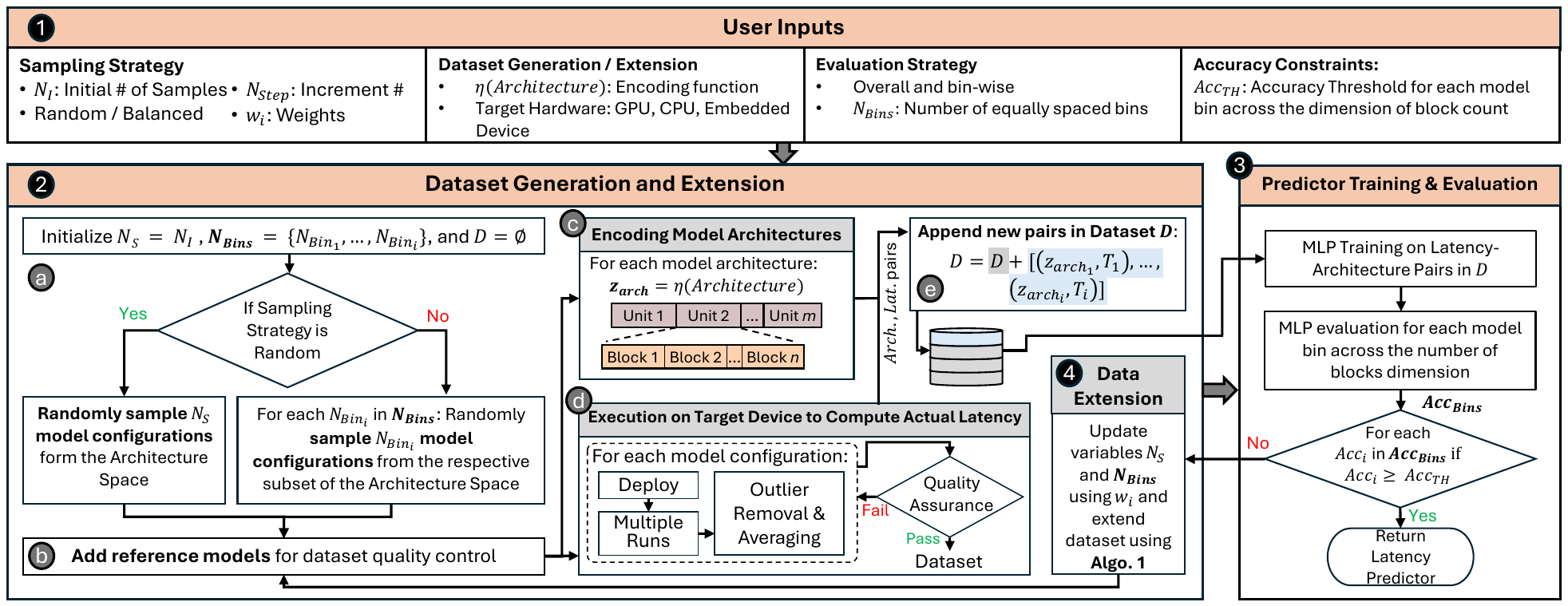}
    \vspace{-20pt}
    \caption{Overview of our ESM Framework.\vspace{-30pt}}
    \label{fig:methodology}
\end{figure*}
\subsection{Overview}
Fig.~\ref{fig:methodology} provides an overview of our novel framework that can produce an accurate latency predictor. The user defines number of initial samples, sampling strategy, encoding type, evaluation strategy, and accuracy constraints. Based on the selected sampling strategy, model configurations are sampled from the Architecture Space. The selected architectures are encoded, and also have their inference performed on the target hardware. These architecture-latency pairs make up the dataset. An MLP-based latency predictor is trained on this dataset, and the performance of the predictor is evaluated. If the user-defined accuracy constraints are met, the framework returns the \textcolor{black}{latency predictor}. Otherwise, the dataset is extended and the predictor is retrained until the accuracy constraints are met. 

\subsection{User Inputs}
The user defines the following inputs for the framework (see \circled{1} in Fig.~\ref{fig:methodology}):
\begin{enumerate}
    \item \textbf{Sampling strategy:} $random$ or $balanced$. 
    \item \textbf{Target Hardware:} GPU, CPU or an embedded device.
    \item $\textbf{N}_\textbf{I}$: Number of initial samples to be sampled from the architecture space.
    \item $\textbf{N}_{\text{\textbf{Step}}}$: Number of additional samples that will be added to $N_I$ during dataset extension. 
    \item $\textbf{w}_\textbf{i}$: Weights assigned to below accuracy and above accuracy threshold bins.
    \item $\eta{\text{\textbf{(Architecture)}}}$: Architecture encoding function
    \item \textbf{Evaluation Strategy}: Overall or bin-wise accuracy
    \item $\textbf{N}_{\text{\textbf{Bins}}}$: Number of equally spaced bins along number of blocks (depth) dimension on which the dataset will be divided
    \item $\textbf{Acc}_{\text{\textbf{TH}}}$: Accuracy threshold for $N_{\text{Bins}}$ to meet to pass the evaluation strategy
\end{enumerate}

\subsection{Dataset Generation}
Once the user has defined their inputs, the framework proceeds based on the selected sampling ($random$ or $balanced$) strategy (see \circled{a} in Fig.~\ref{fig:methodology}). Moreover, based on the number of blocks in each model, the dataset is divided into equally spaced $N_{Bins}$ bins. 
\begin{enumerate}
    \item \textbf{Random Sampling:} 
    In this sampling strategy, we randomly select a given number of models from the entire architectural space. 
    Fig~\ref{fig:encoding} outlines the general structure of our models. Each model is made up of $k$ units; a unit is composed of $l$ number of blocks. For each block, user defines features that they want to vary. For example, in our experiments we fixed each ResNet model to have 4 units, and each unit could have 1 to 7 blocks. When randomly selecting for number of blocks of each unit, a Gaussian spread between the least and the most number of blocks is seen due to the central limit theorem. This bias in the sampling distribution forces the predictor to perform poorly for models present in the corner cases.  
    \item \textbf{Balanced Sampling:} To counter the bias that arises naturally with random sampling, the user can select $balanced$ strategy. The entire architecture space is divided into equally spaced bins, $N_{\text{Bins}}$, based on the number of blocks. For each bin in $N_{Bins}$, equal number of models are sampled from the architecture space. This prevents the under-representation of the corner cases and allows the predictor to effectively learn their trends as well. Each block's kernel size and width expansion ratio is still randomly selected. \par 
Once models have been sampled from the architecture space, reference models (randomly selected from the architecture space) are added to the sample space. These reference models will be later used for dataset quality control.

    \item \textbf{Execution on Target Device and Latency Computation:} On the user-defined target hardware, inference is performed (see \circled{d} in Fig.~\ref{fig:methodology}). For every sampled model, we record the latency 150 times. The slowest and the fastest 20\% of the inferences are discarded, and average is computed of the remaining 60\%. As a quality control measure for our dataset, the latencies of the reference models are analyzed. Using reference models, we ensure the quality of the dataset. This enables us to record the  variance of our dataset, and identify outliers. If the variance is less than 3\% (Fig.~\ref{fig:variance}), the dataset passes the quality assurance; otherwise, it fails and the execution is performed again.
    \begin{figure}
        \centering
        \includegraphics[width=0.8\linewidth]{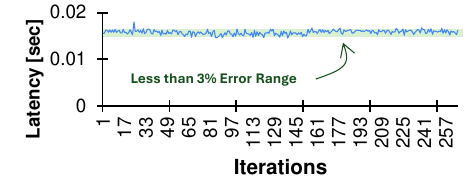}
        \vspace{-10pt}
        \caption{Variance Plot of Reference models. Most of reference model instances fall within the 3\% error boundary. The outliers are removed to improve the dataset quality.\vspace{-15pt}}
        \label{fig:variance}
    \end{figure}
    \begin{figure*}
        \centering
        \includegraphics[width=1\linewidth]{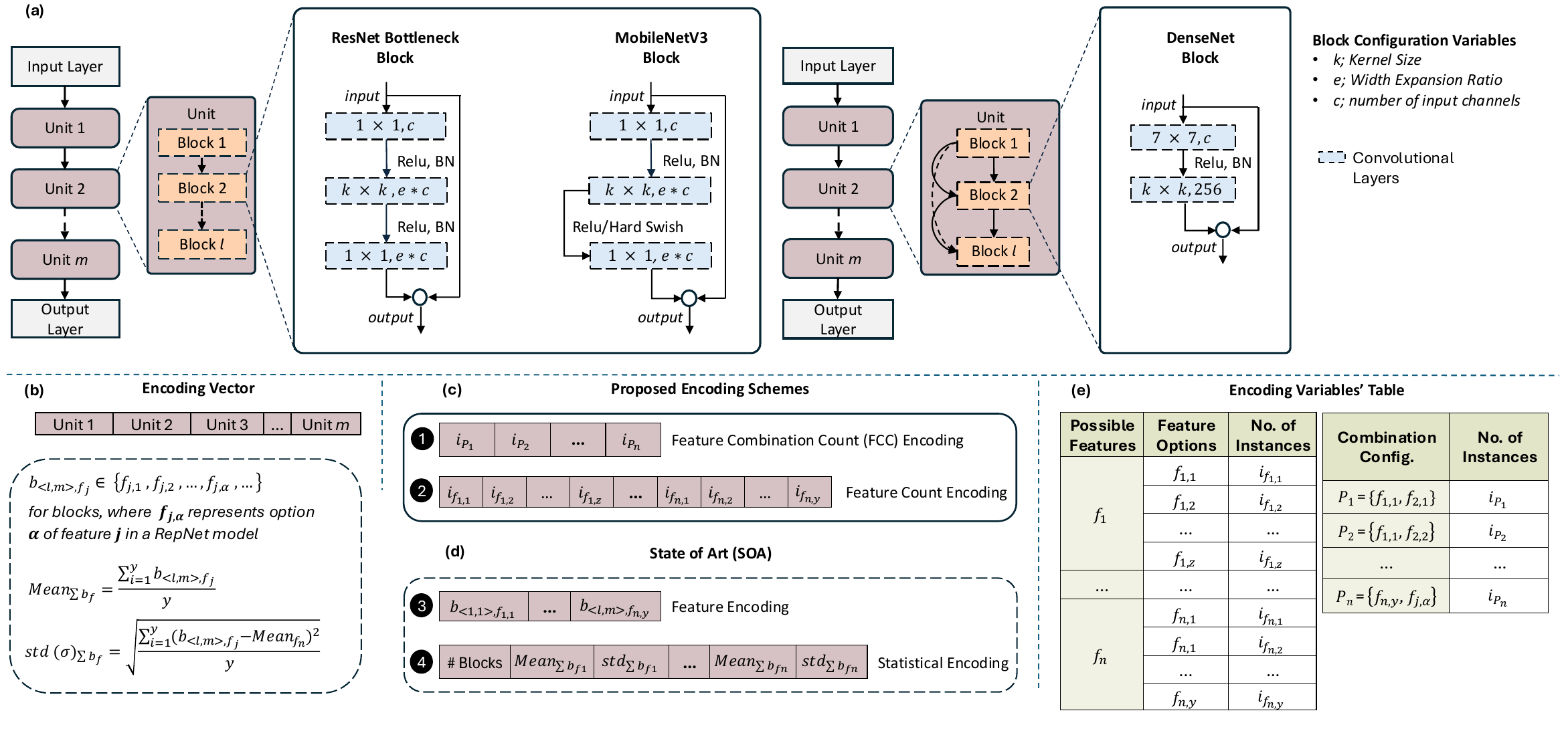}
        \vspace{-25pt}
        \caption{(a) General representation of ResNet, MobileNetV3, and DenseNet supernets. (b) General representation of encoding, illustrating concatenation of unit-level features. (c) Our proposed encoding schemes. (d) State-of-the-art encoding schemes (e) Table containing information of variables used in the encoding schemes.\vspace{-20pt}}
        \label{fig:encoding}
    \end{figure*}
    \item \textbf{Encoding Sampled Model Architectures:} Architectures of DNNs can be represented in various forms. Model architecture configurations after \circled{b} in Fig.~\ref{fig:methodology}, get encoded into vectors (see \circled{c} in Fig.~\ref{fig:methodology}). Let's say that a specific encoding scheme ($\eta$) is a function that translates an architecture to a vector. \[\textbf{z}_{\text{\textbf{arch}}}=\eta{\text{(Architecture)}}\]
    These representations depend largely on the type of architectural space the user defines. Typically, one hot vectors can be utilized to reflect any DNN. But such a representation gives rise to vectors that are extremely lengthy and sparse. Several state-of-the-art techniques employ feature encoding to shorten and reduce the sparsity of the vectors while aiming to maintain feature information~\cite{hassantabar2022curious}\cite{li2022inference}\cite{mendoza2020predicting}. Another common technique is to take mean and standard deviation of features~\cite{wang2020hat} (Fig.~\ref{fig:encoding}d). Although both of these embeddings significantly shorten the vector as compared to one-hot vectors, they do not completely solve the problem. Statistical encoding (Fig.~\ref{fig:encoding}d) make the vectors too dense, causing loss of crucial information, and disabling the predictor to efficiently learn all parameters. On the other hand, Feature encoding (Fig.~\ref{fig:encoding}d) \textcolor{black}{does} detail out features of an architecture, the vector is still quite lengthy and sparse. \par
    We propose an encoding scheme, Feature Combination Count (FCC) Encoding, that is not only significantly shorter than one-hot vectors, but also is not too dense (Fig.~\ref{fig:encoding}c). FCC first makes combinations of all possible features of a block, and then counts the number of instances each combination is used in the given model. This allows the vectors to retain enough information about any specific model's features for the predictor to perform optimally, while also getting rid of sparsity. We also experimented with Feature Count Encoding (Fig.~\ref{fig:encoding}c). This embedding counts the number of occurrences of each feature. Granted, this encoding retained a healthy amount of feature information; however, the vectors were still lengthy and sparse.

\end{enumerate}

The dataset $D$ is made up of these architecture-latency pairs obtained (Fig.~\ref{fig:methodology}e). 

\subsection{Predictor Training and Evaluation} 

The latency predictor is an MLP-based regression model designed to predict the latency of any given model architecture within a specific sample space for the target hardware. 
Primarily, we train our MLP on dataset $D$ and perform evaluation on each of the bins (see \circled{3} in Fig.~\ref{fig:methodology}). This evaluation is based on the accuracy threshold, $Acc_{\text{TH}}$, defined by the user. If all the bin's accuracy, $Acc_{\text{i}}$, is greater than the threshold, $Acc_{\text{TH}}$, the predictor meets the evaluation criteria. If not, all the accuracies of the bins, $Acc_{\text{Bins}}$, are noted. The framework proceeds to dataset extension, after which the predictor is retrained and re-evaluated.

\begin{algorithm}[h]
\footnotesize
\caption{Pseudo-code for Dataset Extension}
\KwIn{$SS$: Entire Sample Space}
\KwIn{$SS_{\text{I}}$: Initial Sampling Space}
\KwIn{$N_{\text{Step}}$: Total number of additional training samples to select}
\KwIn{$N_{\text{I}}$: Number of Initial Samples}
\KwIn{$Acc_{\text{TH}}$: Accuracy threshold for bin categorization}
\KwIn{strategy: Sampling strategy, $``random"$ or $``balanced"$}
\KwIn{$w_1 = \text{Weight of Below Threshold Bins}$}
\KwIn{$w_2 = \text{Weight of Above Threshold Bins}$}
\KwOut{Randomly sampled additional data $SS_{\text{I,rand}}$}
\KwOut{Weighted sampled additional data $SS_{\text{I,weighted}}$}

\If{strategy = random}{
    Randomly select $N_{\text{Step}}$ samples from $SS$ to form $SS_{\text{rand}}$\;
    Append $SS_{\text{rand}}$ to $SS_{\text{I}}$\\
    $SS_{\text{I,rand}} \gets SS_{\text{I}} + SS_{\text{rand}}$\\
    \Return $SS_{\text{I,rand}}$\\
}
\ElseIf{strategy = balanced}{
    Initialize empty list of No.of samples per bin: $\textbf{N}_{\text{\textbf{Bins}}}$\\
    \textcolor{gray}{/* Separate bins $N_{\text{Bin}_{\text{i}}}$  into two groups *}/\\
    $Below_{\text{Acc}_{\text{TH}}} \gets$ Bins with accuracy $<$ $Acc_{\text{TH}}$\\
    $Above_{\text{Acc}_{\text{TH}}} \gets$ Bins with accuracy $\geq$ $Acc_{\text{TH}}$\\
    
    \For{$Below_{\text{Acc}_{\text{TH}}}$ and $Above_{\text{Acc}_{\text{TH}}}$}{
    $N_{\text{B}_{\text{TH}}} = \text{len}(Below_{\text{Acc}_{\text{TH}}})$\\
    $N_{\text{A}_{\text{TH}}} = \text{len}(Above_{\text{Acc}_{\text{TH}}})$\\}

    \textcolor{gray}{/* Calculate samples per $N_{\text{Bin}_{\text{i}}}$ */} \\
    $N_{\text{norm}} = w_1 \times N_{\text{B}_{\text{TH}}} + w_2 \times N_{\text{A}_{\text{TH}}}$\\
    $N_{\text{Below}_{\text{Acc}_{\text{TH}}}} \gets N_{\text{Step}} \times \frac{w_1}{N_{\text{norm}}}$\\
    $N_{\text{Above}_{\text{Acc}_{\text{TH}}}} \gets N_{\text{Step}} \times \frac{w_2}{N_{\text{norm}}}$\\
    $N_{\text{Bins}} \gets \lceil N_{\text{Below}_{\text{Acc}_{\text{TH}}}} \rceil + \lceil N_{\text{Above}_{\text{Acc}_{\text{TH}}}} \rceil$\\

    \For{$i$ in $N_{\text{Bin}_i}$}{
    Sample $N_{\text{Bin}_i}$ from $SS$\\
    Append each sampled $N_{\text{Bin}_i}$ to $SS_{\text{weighted}}$\\
    }
    
    Append $SS_{\text{weighted}}$ samples into $SS_{\text{I}}$\\
    $SS_{\text{I,weighted}} \gets SS_{\text{I}} + SS_{\text{weighted}}$\\
    \Return $SS_{\text{I,weighted}}$\;
}
\end{algorithm}

\subsection{Dataset Extension} 
If the predictor fails to meet the evaluation criteria, the framework looks to extend the dataset for re-training and re-evaluation of the predictor (see \circled{4} in Fig.~\ref{fig:methodology}). This extension is executed using \textbf{Algo.~1.} \par
The user defines total number of additional training samples to select, $N_{\text{Step}}$. If the sampling strategy is $random$, $N_{\text{Step}}$ samples are randomly selected from the entire sample space and appended to the initial sample space. These samples then proceed and \textcolor{black}{undergo} the usual dataset generation pipeline where first reference models are added, then the samples are encoded and executed on a target device. These new architecture-latency pairs are appended to the previous dataset $D$. The MLP is then re-trained and re-evaluated. This loop repeats until all the bins' accuracies, $Acc_{\text{i}}$, meet the threshold accuracy, $Acc_{\text{TH}}$. \par
If the user had defined the sampling strategy as $balanced$, a $weighted$ approach with regards to the bins is taken. The bins, $N_{\text{Bins}}$ are separated into two categories: Below Accuracy Threshold ($Below_{\text{Acc}_\text{TH}}$), and Above Accuracy Threshold ($Above_{\text{Acc}_\text{TH}}$). For each category, number of samples are calculated by applying weights $w_1$ and $w_2$, respectively, to $N_{\text{Step}}$ for each category of bins. This can be seen in \textbf{Algo.~1}. This creates a bias in the dataset towards the below accuracy threshold. When re-training, the MLP learns their patterns as well and eventually crosses the accuracy threshold, $Acc_{\text{TH}}$.

\section{Results and Discussion}

\subsection{Experimental Setup}

\begin{table}
    \centering
    \caption{Supernet architectures and associated hyperparameters}
    \label{tab:configs}
    \resizebox{\linewidth}{!}{
    \begin{tabular}{|>{\centering\arraybackslash}m{0.25\linewidth}|>{\centering\arraybackslash}m{0.24\linewidth}|>{\centering\arraybackslash}m{0.22\linewidth}|>{\centering\arraybackslash}m{0.18\linewidth}|} \hline  
         \textbf{Variable}&  \textbf{ResNet}&  \textbf{MobileNetV3}& \textbf{DenseNet}\\ \hline  
         Stage Width List &  [$2^{8}$, $2^{9}$, $2^{10}$, $2^{11}$]& [$2^{4}$, $2^{5}$, $2^{6}$, $2^{7}$]& N/A\\ \hline  
         \# of Units& 4 & 4 & 5\\ \hline 
         \# of Blocks per Unit.  & $\{1,2,...,7\}$ & $\{1,2,...,7\}$ & $\{1,2,...,20\}$ \\ \hline 
         Kernel Size Options & \{3, 5, 7\} & \{3, 5, 7\} & \{1, 3, 5, 7, 9\}* \\ \hline  
         Width-Expansion Ratio Options & \{1/2, 2/3, 1\} & \{1/2, 2/3, 1\} & N/A \\ \hline  
         \# of Archs.  & $8.38 \times 10^{26}$ & $8.38 \times 10^{26}$ &\(10^{10}\) \\ \hline 
    \end{tabular}}
    \vspace{-5pt}
    \begin{flushleft} \scriptsize * For DenseNet, each kernel size is applied to all the blocks in a unit.  \end{flushleft} \vspace{-20pt}
\end{table}

In this section, we describe the supernet architectures, regression models, and target hardware devices used in our experiments. We used three different supernet architectures: ResNet, MobileNetV3, and DenseNet. The associated hyperparameters are presented in Table~\ref{tab:configs}. 

\textbf{Regression Model:} 
We used an MLP-based regression model consisting of three fully-connected layers, each with a hidden dimension size of 64. The model is trained using Mean Squared Error (MSE) as the loss function and the Adam optimizer, with a learning rate of 0.01 and a weight decay factor of \(10^{-4}\). Unless mentioned otherwise, the predictor is trained on 8,000 samples and tested on 4,000 samples. 

\textbf{Target Hardware:} We evaluated our framework on multiple devices presented below:
\begin{itemize}
    \item \textbf{CPUs:} AMD Ryzen Threadripper 5975, Raspberry Pi 4
    \item \textbf{GPUs:} NVIDIA RTX 4090, NVIDIA RTX 3080 Max-Q
\end{itemize}

\subsection{Effectiveness of Proposed Encoding vs SoTA}
 \begin{figure*}
    \centering
    \includegraphics[width=1\linewidth]{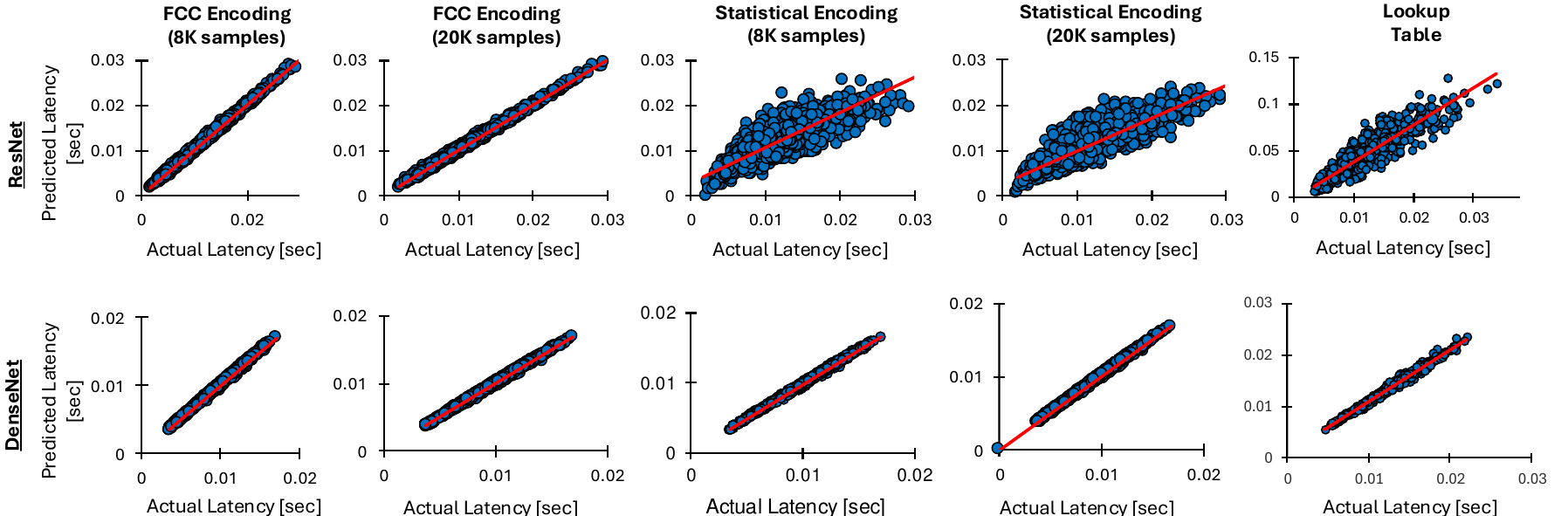}
    \vspace{-20pt}
    \caption{Scatter plots illustrating a comparison between FCC-based MLP with Statistical Encoding-based MLP and Lookup-table for ResNet(Top-row) and DenseNet(Bottom-row). These results are for NVIDIA RTX 4090. \vspace{-10pt}}
    \label{fig:results1A}
\end{figure*}
\begin{figure}
        \centering
        \includegraphics[width=1\linewidth,height=0.5
        \linewidth]{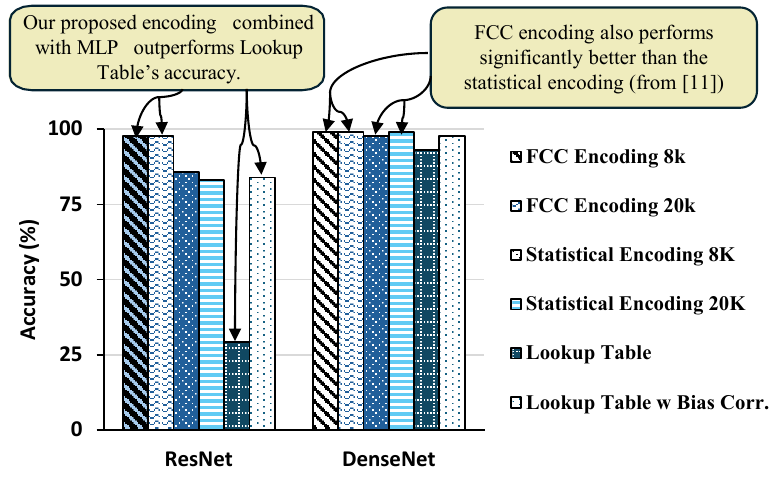}
        \vspace{-20pt}
        \caption{\textcolor{black}{Average accuracies' comparison for the cases presented in Fig~\ref{fig:results1A}.}\vspace{-10pt}}
        \label{fig:1B}
    \end{figure}
    Fig.~\ref{fig:results1A} shows scatter plots of predicted latencies against actual latencies for ResNet and DenseNet. For this experiment, we trained the latency predictor on 8,000 and 20,000 samples using different encoding schemes for ResNet and DenseNet super-nets. The plots show a comparison of the latency predictor's performance using different encoding schemes (proposed FCC vs. Statistical), and a different (Look-up Table-based) surrogate model. These results are further summarized in Fig.~\ref{fig:1B} where it can be seen that the MLP trained on FCC encoding comprehensively outperforms Look-up Table and Statistical Encoding methods. The predictor trained for ResNet supernet on FCC encoded data gave average accuracies of 97.6\%, and 97.8\% for 8,000 and 20,000 samples, respectively. Conversely, statistical encoding gave average accuracies of 85.8\% and 83.1\% for 8,000 and  20,000 samples, respectively. \textcolor{black}{Performing bias correction using a linear regression model significantly improved the Lookup Table predictions with an average accuracy of 83.9\% for ResNets. Similarly, FCC gave 99\% accuracy, while Lookup Table with Bias Correction gave 97\% accuracy for DenseNet.}

\subsection{FCC Encoding Effectiveness Across Devices}
\begin{figure}[h]
    \centering
    \includegraphics[width=1\linewidth]{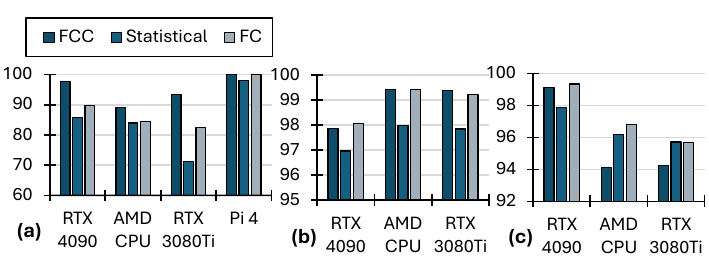}
    \caption{\textcolor{black}{Showcasing effectiveness of the proposed encoding schemes across different target devices: NVIDIA RTX 4090, AMD Ryzen Threadripper 5975, NVIDIA RTX 3080 Max-Q, and Raspberry Pi 4. The three plots are for (a) ResNet, (b) MobileNetV3, and (c) DenseNet.\vspace{-10pt}}} 
    \label{fig:C}
\end{figure}

Figure~\ref{fig:C} displays the performance of the predictor on the different target devices for the DNNs in Fig.~\ref{fig:encoding}\textbf{(a)}. Again, for each device we present a comparison of our proposed \textcolor{black}{FCC and FC encodings} with SoTA Statistical Encoding. Training size samples for RTX 4090 was 8,000, 5,000 for AMD CPU and RTX 3080Ti, and 1,200 for Raspberry Pi 4. It can be seen that for each of the three DNNs, DenseNet, MobileNetV3, and ResNet, \textcolor{black}{FCC and FC outperform} Statistical Encoding on most devices. For ResNet super-net(Fig~\ref{fig:C}(a), the average accuracies across the 4 devices listed were 97\%, 88\%, 93\%, and 99\% for proposed FCC encoding \textcolor{black}{and 90\%, 84\%, 82\% and 99\% for FC encoding}. While the averages for statistical encoding were 85\%, 83\%, 71\%, and 98\%, respectively. For MobileNetV3 super-net (Fig.~\ref{fig:C}(b), the average accuracies across the 3 devices listed were 97\%, 99\%, and 99\% for proposed FCC encoding\textcolor{black}{, and 98\%, 99\% and 99\% for FC encoding,} while the averages for statistical encoding were 96\%, 97\%, and 97\%, respectively. For DenseNet super-net (Fig.~\ref{fig:C}(c), the average accuracies across the 3 devices listed were 99\%, 94\%, and 94\% for proposed FCC encoding \textcolor{black}{and 99\%, 96\% and 95\% for FC encoding,} while the averages for statistical encoding were 91\%, 96\%, and 95\%, respectively. 

\subsection{Random and Balanced Accuracy vs Training Size}
\begin{figure}[h]
        \centering
        \includegraphics[width=1\linewidth, height=0.5\linewidth]{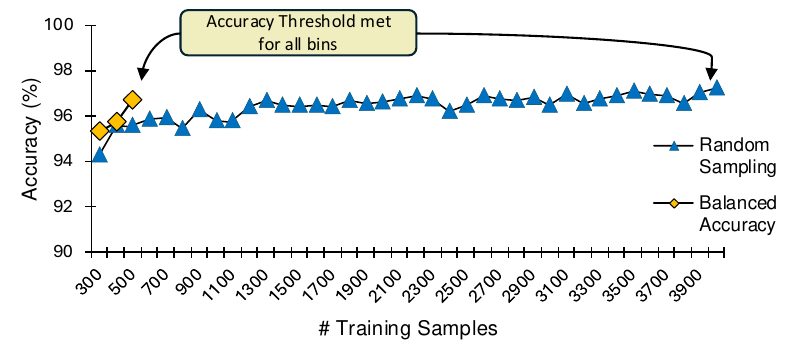}
        \vspace{-10pt}
        \caption{Comparison between Random and Balanced Sampling Strategy for Latency Predictor tested on ResNet super-net on NVIDIA RTX 4090.\vspace{-10pt}}
        \label{fig:D}
    \end{figure}

Figure~\ref{fig:D} shows latency prediction results for $random$ and $balanced$ sampling strategies. For this experiment we computed latency of a ResNet super-net on NVIDIA RTX 4090. We defined $N_I$ as 300 and set $N_{\text{Step}}$ to be 100 and used ESM methodology for training the predictor. The plot clearly showcases that $balanced$ sampling strategy converges much more earlier than $random$ sampling strategy. $Balanced$ sampling strategy converged after 3 iterations with 500 samples, while $random$ sampling strategy took 37 iteration to converge with 4000 samples. 

\section{Conclusion}
In conclusion, this paper highlights the strengths and weaknesses of various surrogate models along with different encoding schemes. Our proposed framework, ESM, along with our novel encoding scheme, FCC, produce results that outperform current State-of-the-Art surrogate models and encoding schemes. Moreover, we provide clear analysis of how the ESM is scalable to a wide array of devices such as GPUs, CPUs, and embedded devices. Additionally, we also present how FCC can be scaled to various DNNs. Based on the insights gained from the analysis, we present a holistic framework that enables accurate, efficient latency prediction using HW-NAS for a variety of target hardware and neural networks, while adhering to user-defined constraints.

\section*{Acknowledgment}
This work was partially supported by the NYUAD Center for Artificial Intelligence and Robotics (CAIR), funded by Tamkeen under the NYUAD Research Institute Award CG010.

\bibliographystyle{IEEEtran}
\bibliography{biblio}

\end{document}